# A Comparison of Non-stationary, Type-2 and Dual Surface Fuzzy Control

Naisan Benatar, Uwe Aickelin and Jonathan M. Garibaldi *Member; IEEE*
School of Computer Science
University of Nottingham
Email: [nxb,uxa,jmg]@cs.nott.ac.uk

*Abstract*—**Type-1 fuzzy logic has frequently been used in control systems. However this method is sometimes shown to be too restrictive and unable to adapt in the presence of uncertainty. In this paper we compare type-1 fuzzy control with several other fuzzy approaches under a range of uncertain conditions. Interval type-2 and non-stationary fuzzy controllers are compared, along with 'dual surface' type-2 control, named due to utilising both the lower and upper values produced from standard interval type-2 systems. We tune a type-1 controller, then derive the membership functions and footprints of uncertainty from the type-1 system and evaluate them using a simulated autonomous sailing problem with varying amounts of environmental uncertainty. We show that while these more sophisticated controllers can produce better performance than the type-1 controller, this is not guaranteed and that selection of Footprint of Uncertainty (FOU) size has a large effect on this relative performance.**

*Keywords: Interval Type-2 Fuzzy, Robot Boat control, Fuzzy Control, Non-stationary*

## I. INTRODUCTION

Fuzzy controllers use the principles of fuzzy sets and fuzzy logic to automate system controllers. The underpinning technique of fuzzy logic was originally introduced by Zadeh in his seminal paper [1]. In this paper, various types of fuzzy set are used including: type-0 which are identical to crisp sets; type-1 where membership is a continuous real value in [0, 1]; and type-2 in which the membership values are themselves type-1 fuzzy sets. The complexity of the sets increases from type-0 to type-2 as the number of dimensions is increased accordingly. Type-1 fuzzy logic has been applied extensively to a range of real-world problems due to the ease with which it can be applied. It has been applied successfully to areas including robotic control, fuzzy decision making and fuzzy classifier systems. However, a number of issues are known to exist in the application of type-1 in problem domains which require decision making in the presence of uncertainty [2]. It is suggested in [3] that such type-1 fuzzy systems have no capacity for modelling uncertainty, which limits their application to more complex real-world problems.

Type-2 fuzzy logic has been proposed as a solution to overcome some of the limitations experienced when using type-1 systems. In contrast to type-1 systems, type-2 systems contain membership functions which in themselves are type-1 fuzzy sets. This adds an extra layer of dimensionality to the system which is postulated to assist in the process of coping with uncertainty in the problem domain. However, this addition of an additional dimension is not without its problems. As a result, additional computational overhead is incurred when implementing type-2 control, which has limited the applicability of type-2 systems to the classical problems solved by type-1 systems, including robotic controllers. This has become less of a problem due to advances in both computational processing power and available memory in computational systems. However, it may still pose a problem when applying type-2 systems to lightweight embedded systems. A more complete overview of interval type-2 fuzzy logic can be found in both [4] and [5].

It has been shown, for example in [6], that type-2 systems can outperform type-1, potentially due to the fact that type-2 fuzzy sets have this 'extra' dimension. There has been some investigation into the reasons underpinning the improved performance of type-2 in comparison with type-1 systems. However, such investigations have been limited in scope. It is also uncertain if improved tuning of type-2 sets or alternative approaches might provide equally good performance without the computational overhead experienced in the application of a full type-2 system.

In conventional type-2 controllers, two control surfaces are obtained, one from the lower bound of interval type-2 defuzzification and one from the upper bound. Due to memory constraints or performance constraints, it is common to implement type-2 fuzzy controllers by calculating the control surface offline, by simply taking the mean of the lower and upper bound value returned by the controller, and then by downloading the resulting surface as a look-up table [7]. In this paper, Birkin and Garibaldi outlined a novel formulation of a controller which maintains the lower and upper control surfaces separately, and then switches dynamically between the surfaces. If the average of the lower and upper control surfaces is used, this novel controller reduces to a 'conventional' interval type-2 controller. Other dynamic combinations are also possible, such that some of the additional information available from an interval type-2 system is maintained in order to be utilised by the controller. This is termed a 'dual-surface' (interval) type-2 controller.

Non-stationary (NS) fuzzy sets, as described in [8] and [9], have been proposed to model variability in human decision making, and may offer a method to alleviate some of the issues raised by the application of type-2 sets. NS fuzzy sets are a relatively new development in the field and are based on the principles of type-1 systems. The 'non-stationary' component refers to variations made to the membership functions defined in a type-1 system. A NS fuzzy set uses perturbations of standard T1 membership functions to produce several slightly different membership function on each iteration through the system. These can then be processed iteratively in the inference system and outputs aggregated, such as by majority vote or mean. NS fuzzy sets have the potential to cope with uncertainty in a problem domain while limiting the additional computational overhead. However, the current extent of comparisons between the different fuzzy controllers is somewhat limited.

It is clear that both theoretical and practical comparisons must be performed in order to understand the differences between the three different fuzzy systems. In this paper we use a simulated autonomous sailing problem (SASP) to examine the practical differences between the different fuzzy control mechanisms. The SASP forms an ideal test-bed for a number of reasons. Firstly, control of an autonomous robot is a challenging problem containing decision making in an uncertain environment. Secondly, this problem involves the interplay between a simulated boat and a dynamic environment, in which data from both the boat and environment are inherently noisy. The amount of uncertainty present in the SASP is an important factor in this application selection. Fuzzy controllers have been researched within the domain of autonomous sailing, especially the application of type-1 controllers, including [10]. In particular, [11] used a type-1 fuzzy system to control an autonomous boat of 1m in length around a predefined course using an attached state machine to handle situations such as tacking (upwind navigation through side to side movement). In this paper it is shown that while the fuzzy controller performed well downwind, it struggled to perform in upwind/tacking scenarios. Fuzzy controllers are not the only method used in the control of autonomous sailing boats. PID controllers [12], and neuro-endocrine approaches [13] have also been successfully applied to navigation problems. Proportional Integral (PI) controllers are also indicated as suitable control mechanisms for autonomous sail boat navigation.

The aim of this paper is to provide a comparison between a number of alternative controllers, including a Proportional-Integral (PI) controller, a type-1 and a conventional interval type-2 controller, when applied to the SASP. We investigate changes in the behaviour of the controllers upon the application of environmental noise, in our case, changes in simulated wind direction. We use a simulator termed `Tracksail' to perform our experiments (see Section III-A). Performing the experiments in simulation has the distinct advantage that experimental set-ups can be precisely replayed and repeated, allowing for the comprehensive analysis of the different techniques. We examine differences between controllers through the resultant control surfaces and by statistical analysis of the experimental results.

This paper is organised as follows: Section **II** describes the components of the fuzzy systems that are under test along with a PI controller that was used as a control. Section **III** outlines the environment and experimental set-up used to make the comparison. This is followed by Section V which presents the results obtained, followed by Section VI which discusses the implications of these results and draws some conclusions. Finally, Section VII outlines some avenues for potential future work.

## II. FUZZY ROBOT CONTROL

*A. Design Decisions*

Stelzer's work on fuzzy sailboats [14], [11] is used as a basis for our controller design. Several changes were necessary due to lack of data about our boat model which were used in the paper to determine some of the parameters, such as *rate of turn* and *heeling* of the boat. Error was retained as the difference between current and desired directions in degrees as shown in Equation 1 while an additional input, change in error has been added to the system. This is defined as the change in the error since the previous iteration of the controller as shown in Equation 2.

$$error = Desired\ Direction - Current\ Direction \qquad (1)$$

$$\Delta error = Current\ Error - Previous\ Error \qquad (2)$$

Each input has five associated fuzzy sets which gives a rule base of 25 rules as shown in Table II-C. The membership functions (MFs) for the terms of the two inputs are shown in Figure 1(a), along with the output of the system which is the percentage change of the rudder of the boat characterised by singleton outputs, shown in Figure 1(f).

Horizontal perturbations of the type-1 membership functions was used to generate our non-stationary controller membership functions. The perturbation function was a horizontal movement defined by a gaussian distribution with a mean of zero. The standard deviation of the distribution was altered several times per batch of experiments to generate several different non-stationary controllers and resulted in input membership functions as shown in Figures 1(b) and 1(d). During execution, each controller would select 30 values from the above distribution to create 30 membership functions which were then processes as a standard type-1 system, the mean of the outputs from each of the 30 systems was taken to give a final output.

Interval type-2 systems in which secondary membership functions are binary instead of continuous can be visualised as a two dimensional area known as the Footprint of Uncertainty (FOU). This makes interval type-2 systems considerably more manageable than the general type-2 variety. We have derived footprints of uncertainty by starting with the simple type-1 and moving a uniform distance along the x-axis in both directions providing a lower and upper bound. This gives FOUs that are very similar in shape to the non-stationary membership functions as shown in Figures 1(c) and 1(e).

A dual surface type-2 controller is implemented to determine if improved results can be achieved through incorporating extra information, such as the upper and lower outputs as outlined in [7]. This employs the algorithm described in Figure 2 for selection of control surfaces and determination of output. This algorithm compares a user chosen metric, in this case the magnitude of the input error with a threshold value.

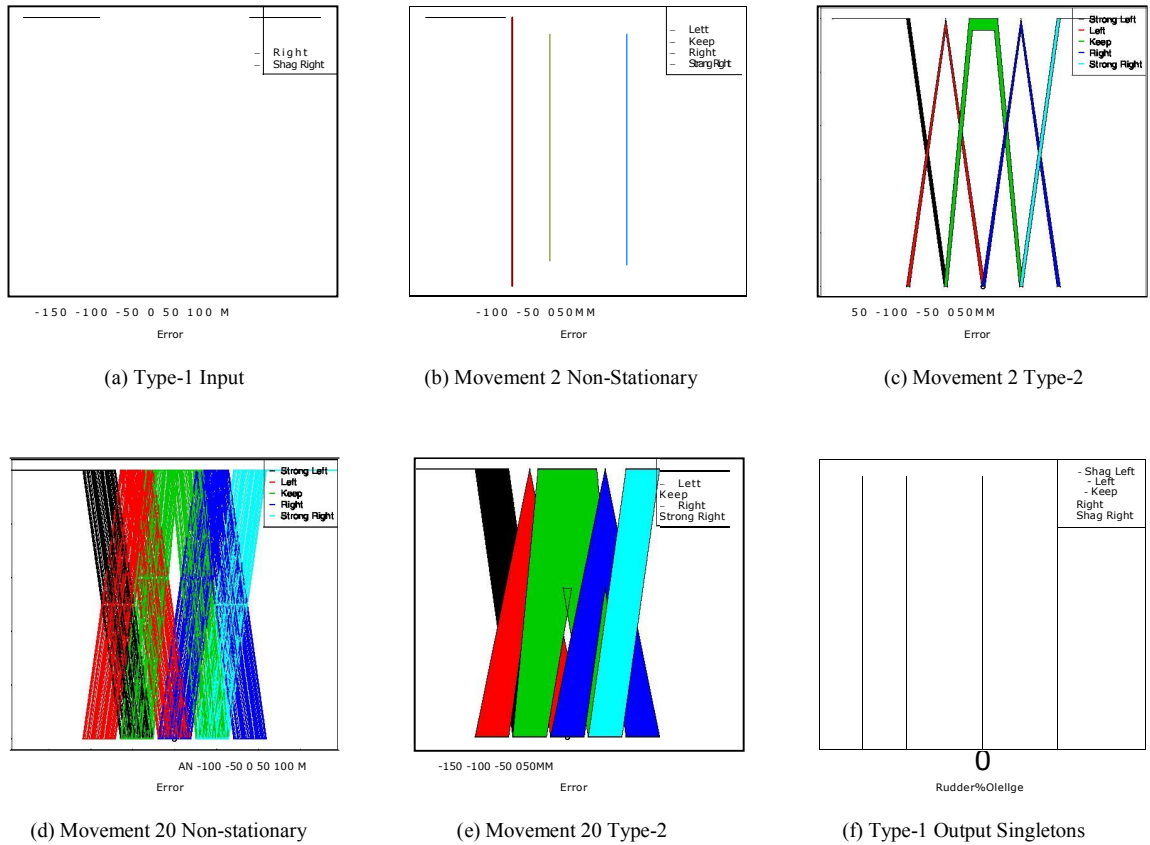

Fig. 1. Membership functions of Type-1, Type-2 and Non-stationary controllers. Unless stated these are input membership functions

```
error = control_var - set_point
diff = abs(error)
if ( diff < THRESHOLD )
   control_action = (LS + US) / 2
else
   if ( error > 0 )
     control_action =
   LS else
     control_action =
US endif
```

Fig. 2. The dual-surface control algorithm

On this basis, the final output of the system is selected from either the lower, upper or mean value. For this comparison, the original method of using the magnitude of the error in the system is retained. As with other controllers, several different threshold values are used to determine any observable effect on the system and its performance.

*B. Tuning & Optimisation*

While manual tuning of the original membership functions ([14] and [15]) was performed we did not specifically perform any automatic tuning for two main reasons. Firstly the intention was to ensure generality and this would not be possible were we to use a specific training data set. Secondly it would be difficult to ensure the same amount of tuning was performed on each different variety of fuzzy due to the different methods required to tune for example, a type-1 system compared to a type-2, which could potentially lead to unfair comparisons being made.

*C. Tacking Behaviours*

With respect to the application of our controllers to SASP, additional behaviours need to be defined. One drawback of using SASP for testing these controllers is the limits of the wind — no boat can sail directly into the wind. To solve this problem, higher level control was added which modifies the desired direction input to the controller. Sailing into the wind in practice requires tacking behaviour. The design of the courses for our experiments attempts to minimise this effect by setting courses which should avoid the need to tack. The main-sail on the boat is controlled by a simple set of rules that were previously developed in conjunction with the PI controller and have been shown to provide acceptable performance.

TABLE I
RULE TABLE FOR TYPE-1 CONTROLLER: ROWS SHOW *delta error* FUZZY SETS; COLUMNS SHOW *error* FUZZY SETS; SR: STRONG RIGHT; R: RIGHT; K: KEEP; L: LEFT; SL: STRONG LEFT

|  | Large Positive | Positive | None | Negative | Large Negative |
|---|---|---|---|---|---|
| Strong Left | sr | sr | r | r | r |
| Left | k | l | r | k | k |
| Middle | k | l | k | r | k |
| Right | k | k | l | r | k |
| Strong Right | l | l | l | sl | sl |

## III. EXPERIMENTAL SETUP

### A. Tracksail

The simulator used for this research, Tracksail has been used previously for development and testing of autonomous sailing robotic systems [16] including the development and tuning of described PI controller. Tracksail is Java based and communicates with controllers by means of a TCP/IP socket allowing any compatible language to be used to control the boat. Each controller is linked into a common framework which provides tacking logic, sensor readings and motor change routines in addition to any other common code required. This ensures consistency between runs with different controllers.

One aim of this work is to investigate changes in the performance of type-2 controllers as the size of the FOU is varied. This is facilitated by the simulator in that it eliminates many categories of uncertainty not directly relevant to this investigation. This includes sensor noise, where the value returned by the sensor does not truly reflect the physical value that it is measuring, for example.

### B. Track Configuration

Each experiment batch consists of 30 simulator runs where the tested controller will attempt to guide the boat towards a single way-point which was defined as 550 metres west of its starting point, defined by the simulator as the point (0,0). In the first batch of experiments the wind direction was fixed as 30 degrees across the boat at a fixed velocity of 7 m/s. This means that the wind will not introduce any noise in this experiment. However, we still observed minor differences between runs, potentially attributed to the timing of interacting processes and delays caused by socket processing in the software and operating system.

A series of experiments introduced wind related noise into the system at two distinct levels. Noise was defined as a specified number of wind changes per second and by the size of the arc defined by the *Max* and *Min* wind direction parameters specified in the simulator which are specified in degrees. Experiment two (medium noise) used an arc of 20° and a single wind change every five seconds, while experiment three (high noise) used a 30° arc with a wind change every three seconds.

### C. Performance Metrics

Two metrics of performance were collected for all controllers. The first is the total root mean square of the error (RMSE) between the current heading compared with the desired bearing. The second metric is the time taken for the boat to complete the set course which is a straight line distance of 550 'metres'. The correlation between RMSE and time taken is not as trivial as in the case of a wheeled robot as the boat controller must balance deviating from a straight line sufficiently to capture enough wind for forward movement, against the extra time taken to perform this manoeuvre. Higher boat speeds in Tracksail can be reached when the boat is parallel to the wind, with the sail set to 90 degrees. However, if this is performed for the entire duration the boat would not reach its destination. Hence, a balance between speed and keeping on course must be found. This becomes increasingly important as noise is introduced into the wind direction variable.

## IV. CONTROLLER SETUP

For reference and control purposes a tuned PI controller is included in the comparison. This controller uses the same input value (error between current and desired bearings) as well as its integral over time instead of the change in error. It uses a P-gain value of 1.7 and an I-gain value of 0.01. These values are derived in previous experiments performed by Suaze and Neal [17].

The running rate for all controllers was fixed at 1Hz. This is chosen to ensure that the more sophisticated controllers can execute a complete control loop. This low running rate may lead to lower performance than optimal as overshoot may occur with slow running controllers. However, as all controllers were subject to the same restrictions, we believe the comparison to be fair.

Five individual values were chosen for the perturbation function of the non-stationary controller, namely 0, 2, 5, 10 and 20 degrees. These values fall within the limits of plausibility for perturbation of the *error* and *delta error* inputs to the system. This provides insight as to where the true optimal value may occur. These measures will also be used to define the FOUs of the interval type-2 systems where the number refers to the width of the FOU at any flat point $(x = c)$.

Numerous parameters in the interval type-2 system (and hence the dual surface controller) were fixed to ensure consistency with the type-1 system. Each experiment involves the variation of a single parameter. In the standard type-2 case we defined a parameter movement which defines the width of the input FOUs and in the dual surface case we used a variable threshold value with a fixed movement value of 5.

As a validation exercise we ran experiment one (low noise) with type-2, non-stationary and dual surface controllers with the parameters including movement value and threshold (where appropriate) set to zero. This is used to highlight that when upper and lower membership functions are set to be equal, the footprint of uncertainty is reduced down to type-1 sets. This confirmed that the performance is equivalent to that of the standard type-1 system. The small disparity was put down to operations in the system such as floating point arithmetic which are performed in different orders in the type-2 based and non-stationary systems. We are therefore satisfied that our type-2 and NS implementations are valid and correct when compared with the type-1.

TABLE II
RMSE AND TOTAL TIME TAKEN FOR COURSE COMPLETION AT LOW
NOISE LEVELS. MEAN AND STANDARD DEVIATION OF 30 RUNS WITH BEST
IN CATEGORY SHOWN IN ITALIC, BEST OVERALL CONTROLLER IN BOLD
AND VALUES THAT ARE STATISTICALLY DIFFERENT FROM THE TYPE-1
CONTROLLER ARE UNDERLINED. PARAMETER REFERS TO MOVEMENT IN
NS AND IT2 CASES AND THRESHOLD IN THE DS CASE

| Variety | Parameter Value | Mean RMSE | Std. Dev RMSE | Mean Time | Time Std.Dev |
|---|---|---|---|---|---|
| PI | N/A | 18.01 | 0.30 | 146.56 | 2.02 |
| Type 1 | N/A | 16.32 | 0.17 | 140.80 | 0.66 |
| Non Stationary | 2 | 17.03 | 0.64 | 139.96 | 1.16 |
| Non Stationary | 5 | *16.72* | *0.54* | ***139.28*** | ***0.63*** |
| Non Stationary | 10 | 16.99 | 1.14 | 139.59 | 1.41 |
| Non Stationary | 20 | 16.74 | 0.55 | 140.07 | 1.10 |
| Type 2 | 2 | 15.97 | 0.62 | *140.42* | 1.03 |
| Type 2 | 5 | *15.84* | 0.28 | 140.65 | 1.18 |
| Type 2 | 10 | 16.04 | 0.53 | 140.80 | 0.66 |
| Type 2 | 20 | 18.94 | 0.57 | 150.03 | 2.69 |
| Dual Surface | 2 | 19.13 | 0.61 | 153.80 | 1.86 |
| Dual Surface | 5 | 19.34 | 1.35 | 150.57 | 3.35 |
| Dual Surface | 10 | 16.73 | 0.59 | 145.43 | 1.33 |
| Dual Surface | 25 | ***15.80*** | ***0.24*** | 149.10 | 7.22 |
| Dual Surface | 50 | 15.99 | 0.25 | *142.38* | *3.59* |

TABLE III
RMSE AND TOTAL TIME TAKEN FOR COURSE COMPLETION AT MEDIUM
NOISE LEVELS. MEAN AND STANDARD DEVIATION OF 30 RUNS WITH BEST
IN CATEGORY SHOWN IN ITALIC, BEST OVERALL CONTROLLER IN BOLD
AND VALUES THAT ARE STATISTICALLY DIFFERENT FROM THE TYPE-1
CONTROLLER ARE UNDERLINED. PARAMETER REFERS TO MOVEMENT IN
NS AND IT2 CASES AND THRESHOLD IN THE DS CASE

| Variety | Parameter Value | Mean RMSE | Std. Dev RMSE | Mean Time | Time Std.Dev |
|---|---|---|---|---|---|
| PI | N/A | 23.25 | 0.30 | 204.69 | 12.97 |
| Type 1 | N/A | 24.47 | 0.76 | 221.34 | 8.46 |
| Non Stationary | 2 | 22.86 | 1.99 | 160.50 | 9.17 |
| Non Stationary | 5 | 22.21 | 4.11 | 172.53 | 23.17 |
| Non Stationary | 10 | 20.27 | 3.18 | 158.53 | 3.61 |
| Non Stationary | 20 | *21.09* | *2.80* | *161.09* | *9.23* |
| Type 2 | 2 | 25.65 | 1.39 | 189.81 | 11.69 |
| Type 2 | 5 | 20.48 | 3.34 | 178.64 | 20.19 |
| Type 2 | 10 | *19.32* | *1.28* | *168.39* | *11.24* |
| Type 2 | 20 | 26.00 | 5.31 | 186.87 | 5.34 |
| Dual Surface | 2 | 20.59 | 0.96 | 168.62 | 7.85 |
| Dual Surface | 5 | 23.06 | 5.10 | 181.94 | 19.03 |
| Dual Surface | 10 | 22.02 | 0.92 | 173.54 | 12.54 |
| Dual Surface | 25 | 19.75 | 3.84 | *171.27* | *12.66* |
| Dual Surface | 50 | ***18.81*** | ***1.61*** | 174.35 | 18.10 |

A. *Hypothesis*

In our experiments we predict that all controllers will show a reduction in performance as the amount of noise introduced into the environment is increased. We anticipate that the performance decrease shown in type-2 systems will be less in magnitude than that of the type-1 controller due to their ability to deal with uncertainty. We also aim elucidate the influence of the size of the FOU on the overall performance of the type-2 controller. In our experiments we ascertain if controllers with larger FOU values will produce improved performances over those with smaller FOUs under higher environmental noise conditions.

V. RESULTS

From Figure 3, it can be observed that as the noise increases (left to right in the subsigures), the courses increase in deviation. This aligns with the hypothesis that increase in environmental noise will result in increasingly non-linear routes.

The results of experiment one (low noise) are shown in Table V, in which it can be observed that the standard type-2 interval controllers have variations in which the RMSE performance is shown to be significantly better than the type-1 and PI controllers. However, there is a peak in the performance increase which occurs at a FOU size of 10 after which performance reduces and drops to lower than that of the PI.

A similar pattern is observed in the case of the dual surface controller. However, in this case both RMSE and time taken start off lower than even the PI but then improves to show the best performance overall. We hypothesise that if the threshold was further increased performance would once again eventually to

TABLE IV
RMSE AND TOTAL TIME TAKEN FOR COURSE COMPLETION AT HIGH
NOISE LEVELS. MEAN AND STANDARD DEVIATION OF 30 RUNS WITH BEST
IN CATEGORY SHOWN IN ITALIC, BEST OVERALL CONTROLLER IN BOLD
AND VALUES THAT ARE STATISTICALLY DIFFERENT FROM THE TYPE-1
CONTROLLER ARE UNDERLINED. PARAMETER REFERS TO MOVEMENT IN
NS AND IT2 CASES AND THRESHOLD IN THE DS CASE

| Variety | Parameter Value | Mean RMSE | Std. Dev RMSE | Mean Time | Time Std.Dev |
|---|---|---|---|---|---|
| PI | N/A | 25.85 | 0.38 | 157.2 | 1.41 |
| Type 1 | N/A | 27.43 | 0.93 | 153.61 | 3.53 |
| Non Stationary | 2 | 31.22 | 4.55 | 153.83 | 7.37 |
| Non Stationary | 5 | 22.21 | 4.11 | 172.53 | 23.17 |
| Non Stationary | 10 | ***20.27*** | ***3.18*** | 158.53 | 3.61 |
| Non Stationary | 20 | 28.69 | 1.35 | *151.23* | *2.60* |
| Type 2 | 2 | 25.48 | 0.66 | *149.70* | 2.08 |
| Type 2 | 5 | *25.33* | *1.36* | 150.19 | 2.33 |
| Type 2 | 10 | 25.83 | 0.93 | 149.77 | 2.75 |
| Type 2 | 20 | 32.72 | 1.92 | 172.37 | 17.31 |
| Dual Surface | 2 | *24.11* | 1.15 | ***141.09*** | ***5.76*** |
| Dual Surface | 5 | 28.93 | 7.41 | 152.49 | 10.02 |
| Dual Surface | 10 | 29.12 | 8.46 | 151.91 | 12.63 |
| Dual Surface | 25 | 26.09 | 0.84 | 151.26 | 2.56 |
| Dual Surface | 50 | 25.95 | 2.66 | 149.81 | 2.86 |

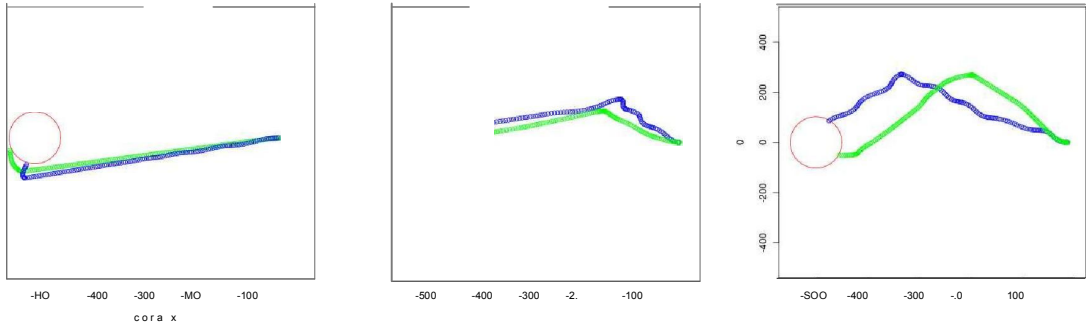

(a) Exp 1 Course Example (Low Noise) (b) Exp 2 Course Example (Medium Noise) (c) Exp 3 Course Example (High Noise)

Fig. 3. Pots of example course taken by PI (green) and Type-2 (blue) at low, medium and high noise levels. Course end point shown as a red circle.

dip below the PI performance level. In this experiment these RMSE improvements do not lead to a significant reduction in time to complete the course compared with the type-1.

Experiment two (medium noise) increases the amount of noise present in the environment with the results shown in Table reftab:Exp2. These clearly show the anticipated drop in performance with the average increase in RMSE being 5.06 and the mean increase in time being 40.2 seconds. A very similar pattern to the previous experiment can be observed in the performance values of the standard type-2 and dual surface results with a peak in performance being observed at a movement level of 10° for the interval type-2 controller, and at a threshold of 50 for the dual-surface controller. Once again it is hypothesised that a threshold greater than that tested would show a drop in performance. Overall, this experiment demonstrates that the type-2 controllers all out-perform the type-1 variety under these noise conditions.

Table V summarises the results of experiment three (high noise). The overall the performance is somewhat lower than for the previous experiment with only one configuration of the dual-surface controller obtaining statistically significant improvements over the type-1 controller, whilst none of the standard interval type-2 or non-stationary approaches achieved this (in the *time taken* metric), though the non-stationary controller did produce two cases in which the RMSE was improved significantly. The mean RMSE increase between experiments one and three were 9.57 with an average time increase of 10.4s.

Two-sided unpaired Mann-Whitney tests are used to determine any statistically significant differences between the type-1 compared against the non-stationary, type-2 and dual-surface controllers. This is performed for both RMSE and *time taken* metrics with a p-value of 0.05 being used to reject the null hypothesis. This test is also performed for the PI and type-1 controllers for all three experiments. The type-1 RMSE proved significantly lower than the PI with low noise (experiment one). For medium and high noise, it was found that the PI is significantly better (lower RMSE) than the type-1 controller. However, in the cases of the total time taken, the type-1 controller was significantly better (lower) than the PI in low and high noise, with the opposite being true for medium noise.

VI. DISCUSSION & CONCLUSIONS

Our findings are summarised below, followed by a more detailed analysis and discussion in the following section.

- At low and high noise levels the more sophisticated controllers generally do not show a statically significant improvement when compared to the type-1 controllers. Specific controllers in each category do show this improvement however.
- At medium noise levels type-2, non-stationary and dual-surface controllers generally do exhibit statistically significant improvements on the type-1 method.
- The results difference between PI and type-1 controllers show that type-1 does improve upon the PI for the RMSE metric and improves under lower noise conditions and low and high noise conditions for the time metric.

The fact that the PI controller actually out-performed the type-1 controller in some experiments, however insignificantly, does tend towards the conclusion that some aspects of the type-1 system were not tuned optimally and that further work may be required in this regard. However, any changes required would also affect the other controllers, which have been based on this type-1. For this reason, we do not anticipate that there would be a great deal of alteration in the general ordering of the performance of the various controllers, were this to be done.

Overall, the performance of the pure type-2 controller was somewhat lower than anticipated and it failed to perform significantly better than the type-1 system in some cases, as is especially evident at low and high noise levels. One potential reason that this could be attributed to is the somewhat naïve method by which the FOUs were generated — that is by blurring a type-1 membership function in an equal distance in either direction. However, it has been shown that simple augmentations made to type-2 systems in form of the dual-surface controller can improve the results of this controller type somewhat and that if alternative methods for selection between the different outputs were investigated performance could be improved even more.

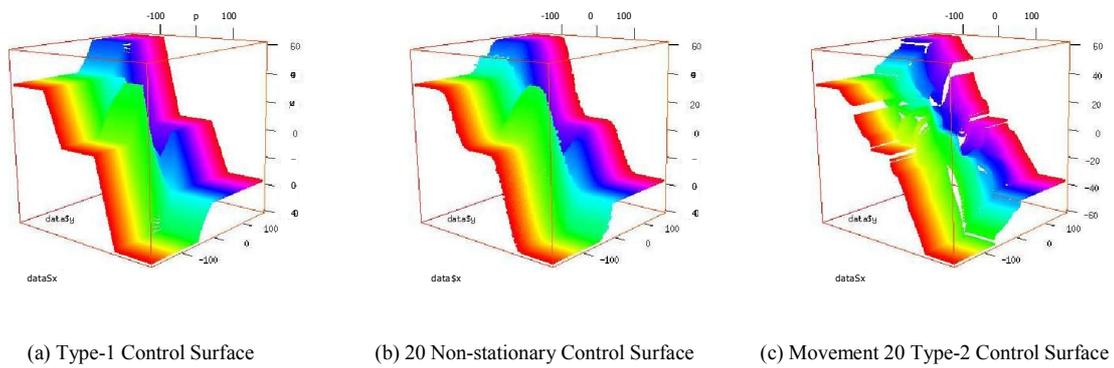

(a) Type-1 Control Surface  (b) 20 Non-stationary Control Surface  (c) Movement 20 Type-2 Control Surface

Fig. 4. Control Surfaces for Type-1, Type-2 (Movement=20) and Non-stationary (Movement = 20) Controllers at varying noise levels

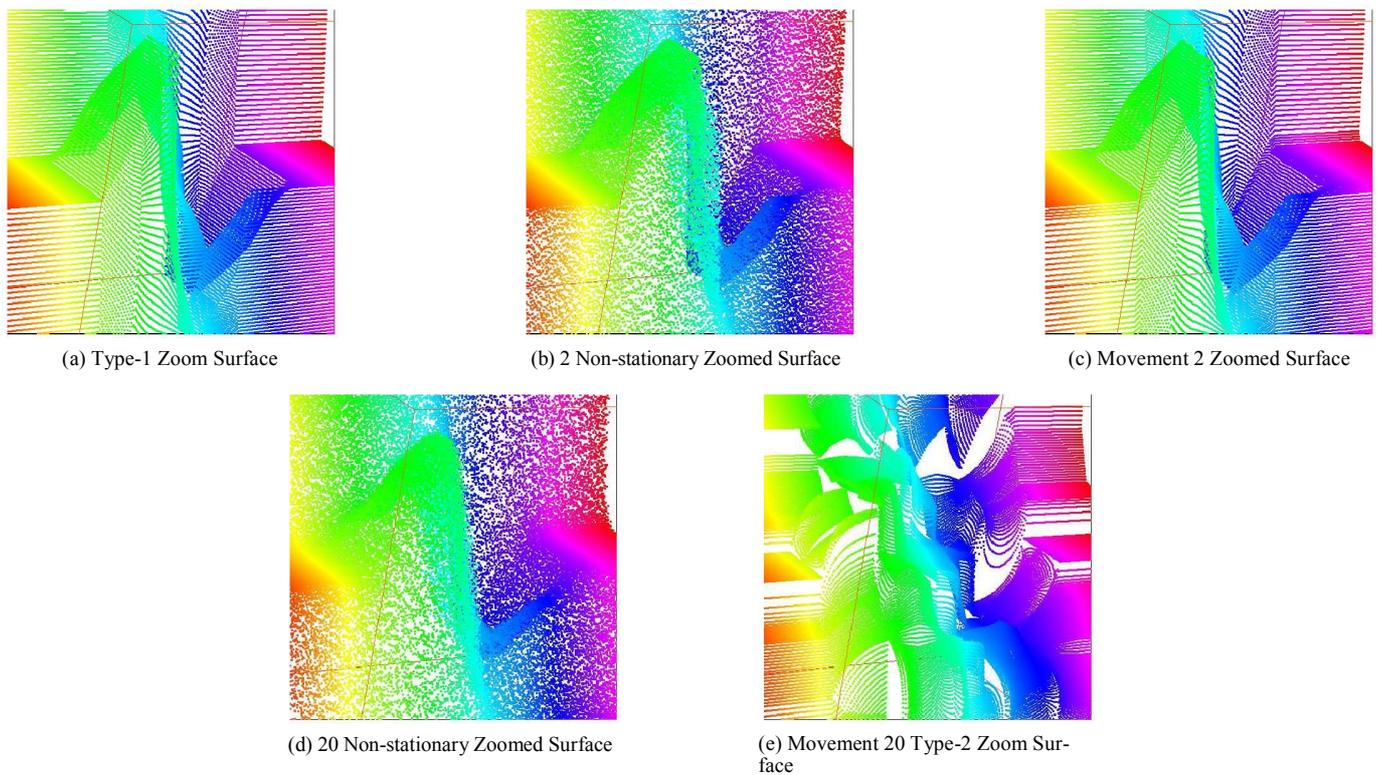

(a) Type-1 Zoom Surface  (b) 2 Non-stationary Zoomed Surface  (c) Movement 2 Zoomed Surface

(d) 20 Non-stationary Zoomed Surface  (e) Movement 20 Type-2 Zoom Surface

Fig. 5. Zoomed view of control Surfaces for Type-1, Type-2 and Non-stationary Controllers at 2 and 20 movement

The results of the non-stationary approach in these experiments shows that in some situations that it can provide equal, or better performance than similar type-2 or dual surface controllers showing that these approaches do offer a viable alternative to type-2 systems as, even though their performance is not always fully comparable, the simplicity to implement in conjunction with the lower computational overhead does give this approach several distinct advantages.

In all of the more sophisticated controllers, the size of the movement (FOU size in type-2 based and horizontal perturbation) had a large effect of the output of the system. In almost all cases there were values selected in which performance decreased markedly. For example, using a value of 20 movement in all three experiments caused the type-2 controller to perform significantly worse than a movement of 10. When the membership functions and FOUs at these levels are observed (Figure 1) it may be suggested that in such cases the degree of overlap would cause considerably more rules to fire than occur at lower movement levels indicating that would would either have to alter the rule-base to accommodate this or change the shape or spacing of the fuzzy sets so that this was less of a problem. This therefore implies two things. Firstly, the shape of the FOU may well be less important that its overall size or width and, secondly, that the selection of this size must be matched to the size of the uncertainty of the environment in which the controller must perform. To support the experimental work done we have also generated control surfaces for each of the described controllers. Inputs were set between -180° and 180° for both inputs and increased in 1° steps which then map to output percentages between -60% and 60%, resulting in Figure 4 and (zoomed in) Figure 5. It can be seen in the type-1 controller (Figure 4(a)) that the gradient changes are very sharply defined (causing sharp controller changes), which contrasts with both the non-stationary and type-2 controllers that are much smoother, leading to smoother controller transitions. Both varieties show a different type of smoothing with the non-stationary displaying more randomly distributed output points and the type-2 showing a more linear smoothing but both do show the same pattern that as the FOU is increased the smoothness of the transition also appears to increase.

While both of these controllers show this smoothing behaviour, it can also be observed in the type-2 surface that there seem to be artefacts in the form of white lines and areas which are not present on the other surfaces and which seem to increase in size as the footprint of uncertainties are increased. The exact nature and reason behind these artefacts are not yet fully understood but have recently been discussed and reasoned about in [18]. However the existence of these discontinuities highlights the complexity of designing and implementing type-2 fuzzy systems and shows that it is far from a trivial task. Currently available software for implementing type-2 systems is also fairly limited — while some free libraries exist many are implemented in MATLAB which is sometimes unavailable due to cost and licensing issues. There are however a great many type-1 libraries available at no charge and adaptation of one of these to create a non-stationary systems is trivial while adding the necessary code to enable type-2 is significantly harder.

In summary, these experiments show that using type-2 control can certainly give performance that exceeds that of type-1 controllers under certain environmental conditions (medium noise, in our case), whereas this performance gain is not necessarily seen in conditions wither of too low noise or too high noise. While this has long been suspected, we believe this paper provides clear experimental confirmation of this observation. We can also see that with the increase of noise this 'optimal' FOU size will increase with it and that the dual-surface controller gives a computationally inexpensive method for getting even more performance out of a type-2 system, but again careful selection of the threshold value is required to achieve these gains.

## VII. FUTURE WORK

We have shown that type-2 based systems can certainly be derived from type-1 and produce performance that exceeds type-1 and PI based approaches under certain conditions, described here as a medium noise environment. Further work is required is to formalise these findings and develop a method by which noise levels can be used to generate an estimate of the FOU required for optimal performance to match the given certain conditions. For example in the above case, we would like be able to estimate an effective FOU size based on the size and frequency of the wind changes in the environment. This method could also be applied to non-stationary approaches to observe its behaviour and performance in these scenarios.

## VIII. ACKNOWLEDGEMENTS

The authors thank the School of Computer Science at the University of Nottingham for supporting this work.